\documentclass{article}

\usepackage[preprint]{neurips_2024}
\usepackage[T1]{fontenc}
\usepackage{amsmath}
\usepackage{graphicx}
\usepackage{microtype}
\usepackage{lipsum}
\usepackage{graphicx}
\usepackage{subcaption} 
\usepackage{booktabs}
\usepackage{multirow}
\usepackage{mdframed}
\usepackage{mdframed}
\usepackage[most]{tcolorbox}
\usepackage{multirow}
\usepackage{amsmath}
\usepackage{bbm}
\usepackage{array}
\usepackage{multirow}
\usepackage{enumitem}
\usepackage{graphicx}
\usepackage{pdflscape}
\usepackage{float}
\usepackage{rotating}
\usepackage{booktabs}
\usepackage{hyperref}
\usepackage[toc,page]{appendix}
\usepackage{booktabs} 
\usepackage{colortbl}
\usepackage{rotating}
\usepackage{amssymb} 
\usepackage{graphicx}

\begin{document}
\title{Leveraging Knowledge Graphs and LLMs for Structured Generation of Misinformation}

\author{
  Sania Nayab$^\diamond$
  Marco Simoni$^{\star\ddagger}$ 
  Giulio Rossolini$^\diamond$
  \\
  $^\diamond${Department of Excellence in Robotics and AI, Scuola Superiore Sant’Anna, Pisa, Italy}
  \\
  $^\star${Institute of Informatics and Telematics, National Research Council of Italy} 
  \\
  $^\ddagger${Sapienza Università di Roma}
}
\maketitle              
\begin{abstract}
The rapid spread of misinformation, further amplified by recent advances in generative AI, poses significant threats to society, impacting public opinion, democratic stability, and national security. Understanding and proactively assessing these threats requires exploring methodologies that enable structured and scalable misinformation generation. In this paper, we propose a novel approach that leverages knowledge graphs (KGs) as structured semantic resources to systematically generate fake triplets. By analyzing the structural properties of KGs, such as the distance between entities and their predicates, we identify plausibly false relationships. These triplets are then used to guide large language models (LLMs) in generating misinformation statements with varying degrees of credibility.
By utilizing structured semantic relationships, our deterministic approach produces misinformation inherently challenging for humans to detect, drawing exclusively upon publicly available KGs (e.g., WikiGraphs).

Additionally, we investigate the effectiveness of LLMs in distinguishing between genuine and artificially generated misinformation. Our analysis highlights significant limitations in current LLM-based detection methods, underscoring the necessity for enhanced detection strategies and a deeper exploration of inherent biases in generative models.

\end{abstract}

\section{Introduction}
\label{intro}
The widespread dissemination of fake information poses significant threats to society \cite{chen2024combating}, particularly with the recent advent of powerful generative AI models (e.g., GPT-4, Stable Diffusion) that significantly enhance the realism of misinformation generation \cite{huang2024creation,huang2023fakegpt}, influencing public opinion and posing risks to national security. The ubiquity of digital media platforms accelerates the spread and impact of misinformation, making it increasingly challenging to detect and mitigate in real-time \cite{wang2024deepfake,wang2024megafake,brau2022minimal}. Consequently, understanding and analyzing how misinformation is generated and propagated has become an essential field of research \cite{shu2020mining}.

The automatic generation of realistic yet deliberately false content is of particular interest, both as a potential threat and as a tool for extensive analysis. On one hand, automating misinformation creation could significantly amplify misinformation attacks, as automated methods can rapidly produce large volumes of credible-sounding false narratives \cite{wang2024megafake}. On the other hand, controlled automated generation can serve as a powerful method to systematically explore, evaluate, and understand the characteristics and vulnerabilities associated with misinformation \cite{huang2023fakegpt}. Thus, developing methodologies that allow structured, scalable, and analyzable misinformation generation is crucial for both defensive strategies and in-depth research into misinformation propagation mechanisms \cite{chen2024combating}.

In this context, we argue that knowledge graphs (KGs) \cite{Bollacker2008FreebaseAC,vrandevcic2014wikidata}, structured repositories that explicitly represent entities and their relationships, offer promising opportunities. KGs inherently encode semantic structures that facilitate the understanding and quantification of the plausibility of potential falsehoods. By leveraging these structures, researchers can automatically derive relational properties between entities within the KG to guide the strategic generation of misinformation, controlling aspects such as relational proximity and perceived credibility. This approach enables the production of sophisticated misinformation that appears credible even under careful scrutiny, complicating both human and algorithmic detection.

\begin{figure}[ht]
    \centering
        \centering
        \includegraphics[width=\linewidth]{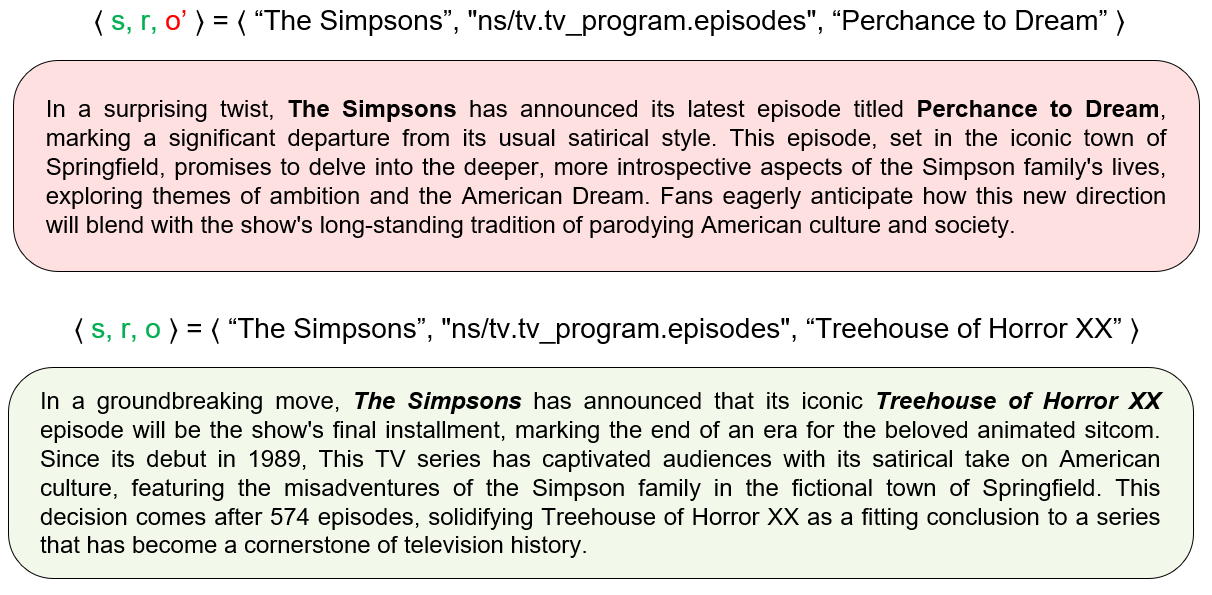}
    \caption{\small{Example of a correct triplet (extracted from the KG, shown in green) and a fake triplet generated by the proposed pipeline (shown in red), along with their corresponding facts generated by the LLM. The fake information (top part in red) represents an incorrect fact, but it is formulated in a coherent and plausible manner.}}
    \label{fig:examples_method}
\end{figure}

Building on this assumption, this paper introduces a structured methodology that exploits predefined KGs to systematically generate templates for realistic fake statements. Our pipeline identifies plausible yet incorrect relational paths within the KG by employing 
its structural properties, enabling the generation of misinformation strategically positioned at multiple levels of plausibility \cite{di2024generating}. As detailed in Section \ref{methodology}, this structured generation leverages the relational and semantic metadata embedded in KGs to ensure control and scalability, while also allowing fine-grained manipulation of stealthiness in relation to potential human scrutiny. To generate the actual misinformation, we provide these fake triplets, categorized by their degree of plausibility, as input prompts to large language models (LLMs). Figure~\ref{fig:examples_method} illustrates an example of fake information generated by the proposed pipeline (in red).

Furthermore, we evaluate the detection capabilities of state-of-the-art LLMs when confronted with these KG-based fabricated statements. Our analysis reveals critical insights into the strengths and limitations of current LLMs in misinformation detection tasks, with implications tied to known challenges such as bias \cite{sallami2024deception,papageorgiou2024survey} and hallucination \cite{wang2023survey}. These findings underscore the need for improved model robustness and a deeper understanding of LLM behavior under varying plausibility conditions.
The main contributions of this work are:
\begin{itemize}
\item The introduction of a structured, knowledge graph-driven approach for generating plausible fake information, exploiting relational metadata to systematically control misinformation plausibility.
\item A comprehensive analysis of generated misinformation across multiple plausibility levels, elucidating the influence of knowledge graph structure on perceived credibility.
\item An evaluative study of LLM capabilities in distinguishing between authentic and generated misinformation, identifying critical detection challenges and biases within current models.
\end{itemize}

\noindent The rest of the document is structured as follows:
we first review relevant prior studies and their limitations in Section~\ref{related}. The methodology, along with the necessary background and notations, is presented in Section~\ref{methodology}. Section~\ref{detection} provides an introduction to the use of LLMs as potential tools for detecting fake information. Experimental results demonstrating the effectiveness of our pipeline are detailed in Section~\ref{exp}. Finally, Section~\ref{con} summarizes the key findings and states potential directions for future research.


\section{Related Works}
\label{related}
\paragraph{Fake news in the LLMs era.}
The generation of false information has become a critical concern, particularly with the advancements in generative AI models that significantly enhance the realism and plausibility of misinformation. Powerful models, such as GPT-4 \cite{achiam2023gpt}, have raised alarms due to their ability to generate high-quality, convincing false content that can easily mislead the public and influence opinions. Huang et al. \cite{huang2023fakegpt} explored the extensive use of prompt engineering with LLMs for fake news generation, highlighting the model's proficiency in generating and explaining fake news.
The generative models, capable of producing human-like text, images, and videos, have potential applications in disinformation campaigns, including elections, social media manipulation, and national security threats \cite{huang2023fakegpt,huang2024creation,wang2024megafake}. 


\paragraph{KGs and LLMs.}
In recent years, several studies have explored the integration of knowledge graphs (KGs) with LLMs. KGs offer a structured and interpretable framework for representing relational and syntactic information, which can guide language generation in a more controlled and semantically grounded manner, potentially reducing issues such as hallucinations and improving factual accuracy.
For example, prior work has investigated the use of KGs to evaluate the performance of LLMs \cite{axelsson2023using,schneider2024comparative}. Other studies \cite{kau2024combining,achiam2023gpt} have demonstrated how models like GPT-3 and GPT-4 can be employed to generate content based on structured data extracted from KGs. Building on these findings, Mauro et al. \cite{di2024generating} show how KGs can be leveraged to identify credible relational links and generate content that is both contextually coherent and structurally aligned with the knowledge base.

Despite these advancements, the use of KGs for generating fake information remains largely underexplored. Given the structured nature of KGs, they hold significant potential for automating the generation of large-scale misinformation with contextual consistency. Moreover, the ability of LLMs to produce highly fluent and plausible, but potentially false, text makes them particularly effective in crafting misinformation that is difficult to detect.

This work brings these two aspects together, proposing an automated framework that generates fake information at scale. The generation process is guided by a novel plausibility metric, allowing the system to produce content with varying degrees of credibility. This enables the creation of misinformation that appears semantically plausible and stealthy, thereby posing new challenges for detection systems.
\paragraph{Detecting Fake Information with LLMs.} As generative models continue to evolve, there is growing interest in exploring the role of LLMs not only in generating, but also in detecting and analyzing false or misleading content \cite{wang2024megafake,wang2024deepfake}. For example, Koka et al. \cite{koka2024evaluating} were among the first to evaluate the effectiveness of various LLMs in fake news detection, comparing the performance of large-scale models such as GPT-4 and Claude 3 Sonnet and smaller models like Gemma 7B.
Despite recent progress, several challenges remain. For instance, a common limitation in the literature is the assumption that the same (often proprietary and unreleased) model is used to both generate and detect fake content, an unrealistic setup for practical deployment. 

To further investigate this, also our work evaluates publicly available LLMs as detectors to assess the quality of the fake content generated through our KG-based pipeline. To ensure a more realistic and reliable evaluation setting, we use LLMs for detection that are distinct from those used to generate the misinformation, and we formalize the detection task as a simple yet effective prompting strategy, enabling an initial assessment of how well off-the-shelf models can identify structured, semantically plausible fake content.

\section{Methodology}
\label{methodology}
{\color{black} 
This section first provides a recap of the background on knowledge graphs (KGs) and outlines the main assumptions related to the definition of fake information in the context addressed by this paper. It then presents a top-down analysis of the proposed approach, beginning with a high-level overview of the steps involved, followed by a detailed description of each step.

\subsection{Background and Assumptions}
\paragraph{Knowledge Graphs (KGs)}. KGs are structured representations of facts and information~\cite{vrandevcic2014wikidata}. Without loss of generality, a fact (or information) $x$ is encoded in the KG as a triplet of the form \( \langle s, r, o \rangle \), where \( s \in \mathcal{E} \) is the subject, \( r \in \mathcal{R} \) is the relation (predicate), and \( o \in \mathcal{E} \) is the object. Here, \( \mathcal{E} \) denotes the global set of entities stored in the KG, and \( \mathcal{R} \) is the set of predicates used to connect those entities as subjects and objects~\cite{luo2023reasoning}. For example, the fact $x =$ \textit{“Paris is the capital of France”} can be represented as the triplet \( \langle s, r, o \rangle = \langle \text{Paris}, \text{capitalOf}, \text{France} \rangle \).

Based on this definition, the complete set of extracted triplets from a knowledge graph is represented as:
\[
T_{\text{KG}} = \{\langle s_1, r_1, o_1 \rangle, \langle s_2, r_2, o_2 \rangle, \dots, \langle s_n, r_n, o_n \rangle\},
\]
where \( n \) denotes the total number of triplets.

\paragraph{Fake Information in the context of KGs.}
Generally speaking, to assess whether a given piece of information $x$ (e.g., a natural language sentence) is correct (real) or incorrect (fake\footnote{please note, we use the terms "incorrect" and "fake" interchangeably, as well as "real" and "correct", while omitting strict formal distinctions.}), we assume the existence of an oracle \( \mathcal{A} \), modeled as a binary classification function. Given a fact \( x \), the oracle assigns a label indicating its truthfulness, such that  
$\mathcal{A}(x) \in \{\textit{real}, \textit{fake}\}.$
Accordingly, an incorrect (fake) fact \( x_\textit{fake} \) satisfies \( \mathcal{A}(x_\textit{fake}) = \textit{fake} \).

Despite the previous formal definition of fake information, in practice, the absence of an oracle (or the impracticality of using one to automate the creation of large sets of potentially fake information) requires a more practical interpretation of the correctness of a fact \( x \) in the specific context we are addressing. To study the generation of correct and incorrect information derived from a KG, we adopt the following assumption: a real fact corresponds to a triple \( t = \langle s, r, o \rangle \) that exists in the KG, such that \( x \rightarrow t \) and \( t \in T_{KG} \).
Analogously, we define an incorrect (i.e., fake) fact \( x_\textit{fake} \) as one that can be mapped to a \textit{fake triple} \( t_\textit{fake} \), i.e., \( x_\textit{fake} \rightarrow t_\textit{fake} \), where we have \( t_\textit{fake} \notin T_{KG} \).
In other words, the core assumption outlined above is that the truthfulness of an information $x$ is determined by the existence of a corresponding triple in the KG, under the assumption that the KG itself is trustworthy.

It is important to note that if we address a KG with a limited number of triples, this may introduce a potential issue: some generated information could be classified as fake according to the previous definition, even though it may not be inherently false. However, in the pipeline proposed next, where the subject and relation are kept fixed, this issue becomes negligible. This is because we only consider subjects and relations that are known to exist in the KG, and we evaluate the truthfulness of information by checking whether the corresponding triple with a given object is missing.

\subsection{Overview of the Pipeline}
\label{m:2}
In this section, we first provide a top-level overview of the proposed pipeline for extracting fake information from a given KG, and then discuss each step in detail.

\begin{figure}[htbp]
    \centering
        \centering
        \includegraphics[width=\linewidth]{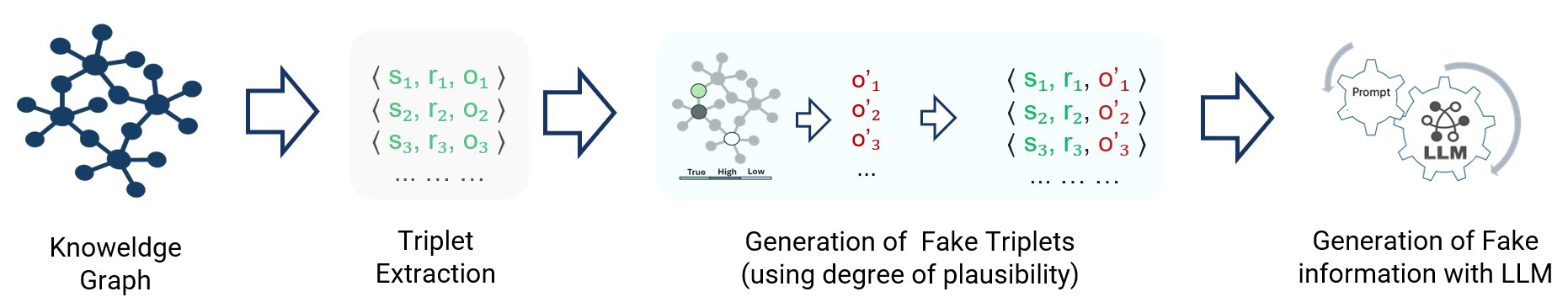}
    \caption{\small{Proposed Methodology for extracting fake information from a given knowledge graph.}}
    \label{fig:method}
\end{figure}

The proposed pipeline is illustrated in Figure~\ref{fig:method}. In short, given a triple \( t \in KG \), the pipeline extracts a fake triple \( t_\textit{fake} \) by modifying the original one, specifically by searching for a new fake object within the KG. It then leverages a LLM to generate the corresponding fake information \( x_\textit{fake} \). For comparison, a related piece of correct information is also shown in green.

In summary, the main steps of the pipeline are as follows:
\begin{enumerate}
    \item \textbf{Triplet Extraction:} Given the KG, all triplets \( t \in T_{KG} \), or a relevant subset, are extracted to serve as reference points for generating associated fake triplets in the next step.
    
    \item \textbf{Generation of Fake Triplets:} For each extracted triplet \( t \), a fake triplet \( t_\textit{fake} \) is generated by exploring the KG in order to replace the original object \( o \) with a different object \( o' \neq o \). In particular, the selection of $o'$ is guided by a degree of plausibility to control the stealthiness of the resulting misinformation (see Section \ref{fakeTriplets}).
    
    \item \textbf{Generation of Fake Information:} Given the fake triplet \( t_\textit{fake} \), a corresponding natural language sentence \( x_\textit{fake} \) is generated using an LLM by adopting a predefined prompt.
\end{enumerate}

\subsection{Generation of Fake Triplets}
\label{fakeTriplets}
Given a triplet \( t = \langle s, r, o \rangle \) from the KG, this step aims to extract a related fake version \( t_\textit{fake} \) by searching for a different object $o'$ in the KG.

\paragraph{Selection of Fake Object Candidates and Degree of Plausibility.}
The extraction of fake information relies on two key points:  
(1) the corresponding fake triplet differs from the original triplet \( t \) only in the object component, i.e., \( t_\textit{fake} = \langle s, r, o' \rangle \), where the subject \( s \) and the relation \( r \) are kept unchanged;  
(2) the exploration of the KG to identify a suitable replacement object \( o' \neq o \) is performed in a way that allows for controlling the degree of plausibility of the resulting fake fact (as detailed below). This enables the study and control of the stealthiness of the fake information generated in the final stage of the pipeline.

As a first step, we restrict the set of candidate objects to those that co-occur with the same relation \( r \) in the KG but are linked to a different subject \( s' \neq s \) (and are never associated with $s$ in any triplet within the KG). 
The idea behind this filtering step is that objects involved in the same predicate \( r \) are more likely to preserve the semantic type or compatibility expected in that relation, thereby increasing the plausibility,  and stealthiness, of the generated fake triplet.
Formally, the filtered candidate set of objects is so defined as:
\[
\mathcal{O}_r = \{ o' \mid o' \neq o \ \wedge\ \exists (s', r, o') \in T_{KG} \wedge\ (s, r, o') \notin T_{KG},\ s' \neq s \}.
\]

To further refine the selection, we rank the candidates in \( \mathcal{O}_r \) based on a plausibility score that captures the structural similarity between the original object \( o \) and the candidate object \( o' \) with respect to their usage in the KG. Specifically, we define a plausibility scoring function \( \mathcal{P}(\cdot) \) as:

\[
\mathcal{P}(o', \langle s, r, o \rangle) = J\big( d(r, o), d(r, o') \big),
\]

\noindent where \( d(r, o) \) denotes the set of distinct subjects \( s \) such that \( (s, r, o) \in T_{KG} \), and \( J \) is the Jaccard similarity. That is, the score compares the sets of subjects associated with \( o \) and \( o' \) under the relation \( r \), favoring candidates that share similar relational connectivity in the KG \cite{di2024generating}.

In other words, this formulation of the plausibility of a fake triplet \( \langle s, r, o' \rangle \) estimates how "realistic" or deceptively correct the triplet appears, based on the similarity in how the original object \( o \) and the candidate object \( o' \) co-occur with the relation \( r \) across the knowledge graph.
This structural similarity suggests that \( o' \) could plausibly replace \( o \) in a sentence without triggering immediate suspicion. 

In the experiments presented in Sections \ref{5.2} and \ref{5.3}, we consider two opposite categories of fake triplets based on their plausibility scores: high-plausibility fake triplets \( o'_{\text{high}} \) and low-plausibility fake triplets \( o'_{\text{low}} \), formally defined as:
\[
o'_{\text{high}} = \arg\max_{o' \in \mathcal{O}_r} \mathcal{P}(o', \langle s, r, o \rangle)
\]
\[
o'_{\text{low}} = \arg\min_{o' \in \mathcal{O}_r} \mathcal{P}(o', \langle s, r, o \rangle)
\]

The objective behind studying, in the experimental part, two levels of triples based on different plausibility levels is the following: fake triples constructed using high-plausibility objects are more semantically similar to the original facts and are therefore more difficult to detect as fake. At the same time, due to their semantic similarity, such fake information may introduce subtler distortions, i.e., misinformation that appears credible but is nonetheless factually incorrect. Conversely, low-plausibility fake triples are expected to be more clearly incorrect and potentially more harmful if mistakenly trusted by users (see the example in Figure \ref{fig:examples_method}, which depicts a low-plausibility case), while also being easier to recognize as fake.

\paragraph{Clarification on the approach used.}
It is important to note that, in theory, starting from a correct reference triplet \( t = \langle s, r, o \rangle \), one could generate a large number of fake triplets by altering the subject \( s \), the relation \( r \), or the object \( o \), resulting in a new triplet \( t_\textit{fake} = \langle s', r', o' \rangle \). However, in the implementation of the pipeline described in the following subsection, we restrict our analysis to the generation of fake triplets by modifying only the object.

The rationale behind this limitation is twofold: first, since the sets of entities used as subjects and objects in the KG are often the same, modifying both \( s \) and \( o \) is unnecessary and computationally inefficient. Second, by restricting modifications to the object only, we can better analyze the plausibility and deception level of the fake information while preserving the original subject and relation. This facilitates direct comparison with the correct version of the fact.

\subsection{Generation of Fake Information Using LLM}

Given a fake triplet $t_\textit{fake}$, extracted in the previous step, the goal here is to generate a natural language fake sentence $x_\textit{fake}$ based on $t_\textit{fake}$.
To achieve this, we use a LLM guided by a predefined prompt based on the fake triplet $t_\textit{fake}$, with the aim of producing a coherent output sentence.
Specifically, we adopted the prompt shown in Figure~\ref{f:genPromot}, which is composed of several parts: a preamble to guide the LLM in the context of sentence generation; a set of additional instructions (described below); the triplet information we aim to target; and finally, the actual request to generate a short news article.

\begin{figure*}[ht]
\begin{tcolorbox}[colback=black!5!white,colframe=black!75!black,
title={Prompt for Generating Fake Information}]
   
    \begin{minipage}{\columnwidth}
    \footnotesize
   You are a professional journalist working for a major entertainment and culture media outlet. 
   Your job is to write short and realistic news blurbs about a wide range of public entities, such as TV shows, movies, albums, books, people, places, or products.  \\
    You are provided with a subject, a brief description, and a triple (subject, predicate, object) in which one fact has been intentionally replaced with incorrect information. 
    Your task is to write a brief but persuasive article that presents the fake information as if it were true. \\
    \end{minipage}  
   
    \begin{minipage}{\columnwidth}
    \footnotesize
    [\textit{Additional rules}] \\
    

    Subject: { \color{blue}\( <s> \)} \\
    Description: {\color{blue}< description of $s$ available from the the KG >} \\
    Fake Triple: {\color{blue}{<\( s \)}, {\( r \)}, {\( o' \)}>} \\

    Write a very short news article (max 3 sentences) based on the triple above. \\
    Use the description to provide background context. 
    \end{minipage}
\end{tcolorbox}%
\caption{\small{{Description of the LLM prompt used to generate fake information based on the original subject $s$, its description from the knowledge graph, the original relation $r$, and the previously computed fake object $o'$.}}}
\label{f:genPromot}
\end{figure*}


The additional writing rules, omitted from the figure for simplicity, are intended to ensure proper formatting and guide the LLM's behavior. These include: (i) use of a confident and informative tone; (ii) do not reveal or hint that the information is incorrect; (iii) use the description as background, without copying it verbatim; and (iv) aim for clarity and plausibility in a compact format. Additionally, we explicitly include a requirement in the final prompt to avoid overly long generations by specifying a limit of no more than three sentences.
Following this, the LLM produces a brief but informative news item based on the given prompt.

It is important to note that the proposed prompt can be applied to any triplet, including a correct one $\langle s, r, o \rangle$, thereby generating a valid factual sentence (as shown in Figure \ref{fig:examples_method}, in green). In fact, this pipeline is also used in the experimental phase to extract correct information, which is then employed to evaluate the ability of the detector to distinguish between true and false content.

\section{Detection of Fake Information Using LLMs}
\label{detection}
To evaluate whether the generated sentences can be automatically detected as fake, we explore the use of LLMs as potential detection tools, leveraging their internal knowledge.
This section introduces a baseline detection approach based on LLMs, which is evaluated and discussed in the experimental sections (\ref{5.2}, \ref{5.3}). There, we highlight the limitations of different models in performing this task and outline interesting challenges.

For each LLM used in the detection task, we utilized a detection prompt (shown in Figure~\ref{fig:DetPrompt}), where we formally ask the model to return a judgment about the correctness of a given fact $x$. 
Please note that here the prompt also includes additional instructions to enforce a consistent output format:
(i) Base your judgment on linguistic cues, world knowledge, plausibility, and consistency;
(ii) Do not explain your answer;

\begin{figure*}[!ht]
\begin{tcolorbox}[colback=black!5!white,colframe=black!75!black,
title={Prompt for Detecting Fake Information}]

    \begin{minipage}{\columnwidth}
    \footnotesize
    You are a fact-checking assistant. Your job is to evaluate whether a given news statement is factual [Real] or misleading [Fake].\\
    Task: Given a news statement, analyze the content and respond with one of the following: \\
    - `[Real]` → if the statement is likely factual or plausible. \\
    - `[Fake]` → if the statement contains false or misleading information. \\
    
    [\textit{Additional writing rules}] \\
    
    \end{minipage}  
   
    \begin{minipage}{\columnwidth}
    \footnotesize
    Evaluate the truthfulness of the following news statement:  {\color{blue}$ < x > $.}\\ 
    Is this statement real or fake?
    \end{minipage}
\end{tcolorbox}%
\caption{\small{Prompt used for evaluating the truthfulness of news statements. The task involves analyzing a given news statement to determine if it is factual (`[Real]`) or misleading (`[Fake]`), based on linguistic cues, world knowledge, and plausibility.}}
\label{fig:DetPrompt}
\end{figure*}


In the experimental section, we evaluate multiple LLMs for the detection task as Judges, and to improve detection reliability. This setup allows us to assess the quality of LLM-based detectability while also quantifying the number of false negatives, particularly in cases where the sentence might closely resemble a true fact.

Moreover, we leverage the use of LLMs as evaluators to assess whether higher plausibility makes it more difficult to automatically detect fake information, thereby supporting the observations made in the previous section regarding the role of plausibility.

\section{Experiments}
\label{exp}
This section presents a set of experiments designed to illustrate the generation of fake facts. In the first part, we showcase examples produced by the pipeline introduced in Section \ref{m:2}, covering both low and high plausibility cases. The goal is to demonstrate how the generated statements, especially those with high plausibility, can effectively mislead human readers.

Subsequently, we investigate the ability of LLMs to detect fake facts generated by our approach. We analyze how detection performance varies with different levels of plausibility and also examine the occurrence of false negatives, where factual information is incorrectly flagged as fake.

\subsection{Experimental Setup}

\paragraph{Data used.} The KG utilized in these experiments is Wikigraphs \cite{wang2021wikigraphs}, a publicly available knowledge graph commonly used to evaluate general knowledge. For our experiments, we focused on categorical knowledge graph triplets, excluding less common categories and those integrated with Freebase when combining Wikipedia data into the graph (such as type, base, etc.). Ultimately, we selected 25 categories for the experiments to ensure a diverse and distinguished generated fake facts and related correct facts for comparisons. In particular, we generated a total of 14450 fake information samples. For comparison, we also included an additional subset of 1540 real facts generated from correct triplets.

\paragraph{Models.} 
We tested two distinct sets of models for fake news generation and detection to ensure a comprehensive and non-overlapping evaluation. For fake news generation, we employed Phi-4 \cite{abdin2024phi}, representing a large-scale language model, and Llama-3-8b \cite{touvron2023llama}, a smaller-scale model. This setup allows us to assess the impact of model size on the realism and plausibility of the generated content.

For fake news detection, we selected a different set of state-of-the-art LLMs: Falcon-40b \cite{almazrouei2023falcon}, Llama-3-70b \cite{touvron2023llama}, and Qwen-2-72b \cite{yang2024qwen2}. 
In Section \ref{5.3} provides a separate analysis of each individual LLM used for detection, enabling a more detailed understanding of their behavior and potential biases when distinguishing between fake and real facts.

All experiments were conducted using the Text Generation Inference (TGI) platform\footnote{\url{https://huggingface.co/docs/text-generation-inference}} and Vllm for Falcon-40b\footnote{\url{https://huggingface.co/tiiuae/falcon-40b?local-app=vllm}} on 8 NVIDIA A100 GPUs.

\paragraph{Detection Metrics.} For all the experiments, we evaluate the capabilities of LLMs (both individually and as a jury) in detecting fake facts. Specifically, this task is formulated as a binary classification problem, where each given fact \( x \) is associated with a ground-truth label \( y \in \{\text{fake}, \text{real}\} \).

Given a set of facts, the accuracy of an LLM in detecting fake or real information is computed using the standard binary accuracy formula:

\[
\text{Detection Accuracy} = \frac{1}{N} \sum_{i=1}^{N} \mathbf{1}\left[D(x) = y_i\right],
\]

\noindent where \( N \) is the total number of evaluated facts, \( y_i \) is the ground-truth label, \( D(x) \) is the prediction provided by the model, and \( \mathbf{1}[\cdot] \) is the indicator function that returns 1 if the condition is true and 0 otherwise.

Please note that, in the following experiments, we do not report confusion matrices or other standard analyses used in machine learning for imbalanced datasets, since the subsets of real and fake facts are examined independently.

\subsection{Qualitative Evaluation of Plausibility in Generated Fake News}
\label{illustration}

\begin{figure*}[!ht]
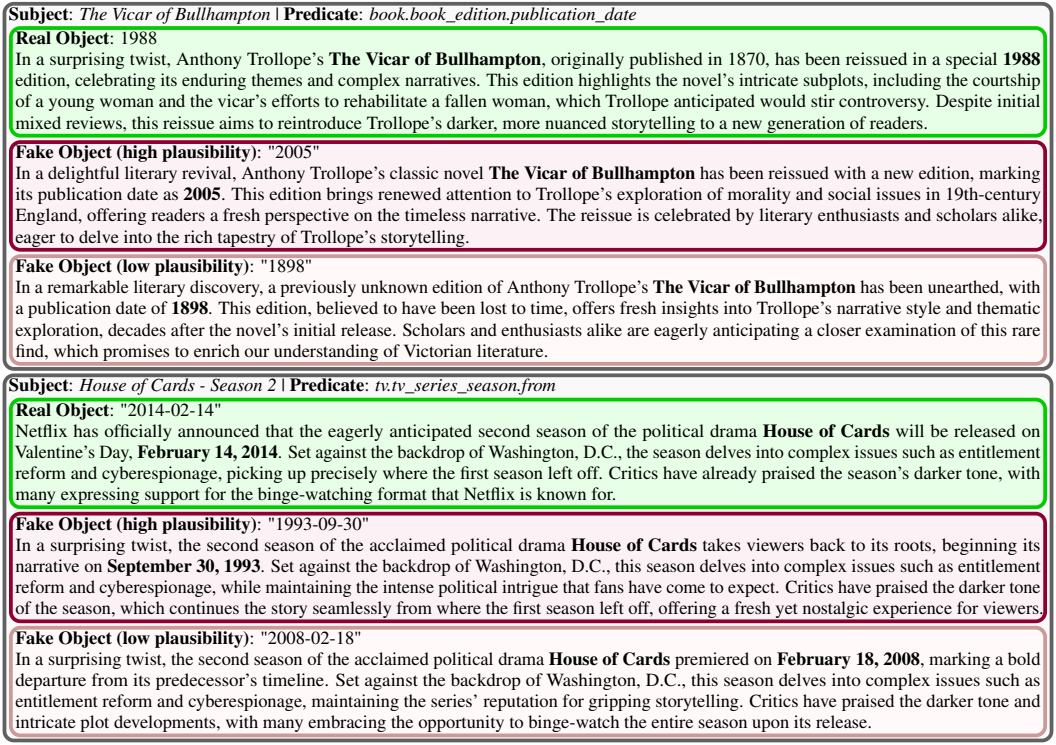

\begin{minipage}{\textwidth}
\begin{tcolorbox}[
    colback=gray!5!white,
    colframe=gray!75!black,
    fonttitle=\scriptsize,
    boxsep=0pt, left=1pt, right=1pt, top=0.5pt, bottom=0.5pt,
    coltitle=black,
    enhanced,
    before skip=0pt, after skip=0pt
]
\scriptsize
\textbf{Subject}: \textit{The Vicar of Bullhampton} |
\textbf{Predicate}: \textit{book.book\_edition.publication\_date}

\begin{tcolorbox}[
    colback=green!10!white,
    colframe=green!80!black,
    fonttitle=\scriptsize,
    boxsep=0pt, left=1pt, right=1pt, top=0.5pt, bottom=0.5pt,
    coltitle=black,
    enhanced,
    before skip=1pt, after skip=1pt
]
\scriptsize
\textbf{Real Object}: 1988 \\
In a surprising twist, Anthony Trollope's \textbf{The Vicar of Bullhampton}, originally published in 1870, has been reissued in a special \textbf{1988} edition, celebrating its enduring themes and complex narratives. This edition highlights the novel's intricate subplots, including the courtship of a young woman and the vicar's efforts to rehabilitate a fallen woman, which Trollope anticipated would stir controversy. Despite initial mixed reviews, this reissue aims to reintroduce Trollope's darker, more nuanced storytelling to a new generation of readers.
\end{tcolorbox}

\begin{tcolorbox}[
    colback=purple!5!white,
    colframe=purple!75!black,
    fonttitle=\scriptsize,
    boxsep=0pt, left=1pt, right=1pt, top=0.5pt, bottom=0.5pt,
    coltitle=black,
    enhanced,
    before skip=1pt, after skip=1pt
]
\scriptsize
\textbf{Fake Object (high plausibility)}: "2005" \\
In a delightful literary revival, Anthony Trollope's classic novel \textbf{The Vicar of Bullhampton} has been reissued with a new edition, marking its publication date as \textbf{2005}. This edition brings renewed attention to Trollope's exploration of morality and social issues in 19th-century England, offering readers a fresh perspective on the timeless narrative. The reissue is celebrated by literary enthusiasts and scholars alike, eager to delve into the rich tapestry of Trollope's storytelling.
\end{tcolorbox}

\begin{tcolorbox}[
    colback=pink!10!white,
    colframe=pink!80!black,
    fonttitle=\scriptsize,
    boxsep=0pt, left=1pt, right=1pt, top=0.5pt, bottom=0.5pt,
    coltitle=black,
    enhanced,
    before skip=1pt, after skip=0pt
]
\scriptsize
\textbf{Fake Object (low plausibility)}: "1898" \\
In a remarkable literary discovery, a previously unknown edition of Anthony Trollope's \textbf{The Vicar of Bullhampton} has been unearthed, with a publication date of \textbf{1898}. This edition, believed to have been lost to time, offers fresh insights into Trollope's narrative style and thematic exploration, decades after the novel's initial release. Scholars and enthusiasts alike are eagerly anticipating a closer examination of this rare find, which promises to enrich our understanding of Victorian literature.
\end{tcolorbox}

\end{tcolorbox}
\end{minipage}

\begin{minipage}{\textwidth}
\begin{tcolorbox}[
    colback=gray!5!white,
    colframe=gray!75!black,
    fonttitle=\scriptsize,
    boxsep=0pt, left=1pt, right=1pt, top=0.5pt, bottom=0.5pt,
    coltitle=black,
    enhanced,
    before skip=0pt, after skip=0pt
]
\scriptsize
\textbf{Subject}: \textit{House of Cards - Season 2} | 
\textbf{Predicate}: \textit{tv.tv\_series\_season.from}

\begin{tcolorbox}[
    colback=green!10!white,
    colframe=green!80!black,
    fonttitle=\scriptsize,
    boxsep=0pt, left=1pt, right=1pt, top=0.5pt, bottom=0.5pt,
    coltitle=black,
    enhanced,
    before skip=1pt, after skip=1pt
]
\scriptsize
\textbf{Real Object}: "2014-02-14" \\
Netflix has officially announced that the eagerly anticipated second season of the political drama \textbf{House of Cards} will be released on Valentine's Day, \textbf{February 14, 2014}. Set against the backdrop of Washington, D.C., the season delves into complex issues such as entitlement reform and cyberespionage, picking up precisely where the first season left off. Critics have already praised the season's darker tone, with many expressing support for the binge-watching format that Netflix is known for.
\end{tcolorbox}

\begin{tcolorbox}[
    colback=purple!5!white,
    colframe=purple!75!black,
    fonttitle=\scriptsize,
    boxsep=0pt, left=1pt, right=1pt, top=0.5pt, bottom=0.5pt,
    coltitle=black,
    enhanced,
    before skip=1pt, after skip=1pt
]
\scriptsize
\textbf{Fake Object (high plausibility)}: "1993-09-30" \\
In a surprising twist, the second season of the acclaimed political drama \textbf{House of Cards} takes viewers back to its roots, beginning its narrative on \textbf{September 30, 1993}. Set against the backdrop of Washington, D.C., this season delves into complex issues such as entitlement reform and cyberespionage, while maintaining the intense political intrigue that fans have come to expect. Critics have praised the darker tone of the season, which continues the story seamlessly from where the first season left off, offering a fresh yet nostalgic experience for viewers.
\end{tcolorbox}

\begin{tcolorbox}[
    colback=pink!10!white,
    colframe=pink!80!black,
    fonttitle=\scriptsize,
    boxsep=0pt, left=1pt, right=1pt, top=0.5pt, bottom=0.5pt,
    coltitle=black,
    enhanced,
    before skip=1pt, after skip=0pt
]
\scriptsize
\textbf{Fake Object (low plausibility)}: "2008-02-18" \\
In a surprising twist, the second season of the acclaimed political drama \textbf{House of Cards} premiered on \textbf{February 18, 2008}, marking a bold departure from its predecessor's timeline. Set against the backdrop of Washington, D.C., this season delves into complex issues such as entitlement reform and cyberespionage, maintaining the series' reputation for gripping storytelling. Critics have praised the darker tone and intricate plot developments, with many embracing the opportunity to binge-watch the entire season upon its release.
\end{tcolorbox}

\end{tcolorbox}
\end{minipage}
\caption{\small{Examples of two fake news generations by Phi-4 and the corresponding correct news (based on the real object available in the KG for each specific triplet). For each fact, two versions of fake news are shown, reflecting low and high plausibility levels.}}
\label{f:illustratedExamples}
\end{figure*}

To provide a qualitative, human-level assessment, without relying on automatic detection mechanisms, we present and discuss two examples of fake news generated using the Phi-4 model.

Figure~\ref{f:illustratedExamples} showcases representative cases where fake information was created by modifying the object component in factual triplets extracted from a knowledge graph, alongside the corresponding real information. For each fact, two manipulated versions are shown: one with high plausibility and one with low plausibility. These examples highlight the difficulty of detecting misinformation when the generated content is semantically coherent and stylistically consistent with genuine text.

In the first case, the subject is The Vicar of Bullhampton, a novel by Anthony Trollope. The real publication date (1988 edition) is replaced with either "2005" (high plausibility) or "1898" (low plausibility). Both fake versions preserve a literary tone and thematic references that align with Trollope’s work and historical context. The "2005" version is particularly deceptive, as modern reprints of classical literature are common and seldom questioned, making the change seem natural and easily overlooked. In contrast, the "1898" version, while still plausible at a surface level, may raise suspicion due to its claim of a “previously unknown edition,” which introduces a higher degree of narrative risk and slightly undermines its credibility.

The second example focuses on House of Cards – Season 2, where the actual release date ("2014-02-14") is altered to either "1993-09-30" (high plausibility) or "2008-02-18" (low plausibility). Both fake versions maintain the stylistic tone and thematic framing of the original. However, the 1993 date appears especially convincing, as House of Cards was originally a British miniseries aired in the early 1990s. 
By contrast, the 2008 version, although chronologically closer to the real targeted release, lacks a meaningful historical or narrative anchor. As a result, it appears more arbitrary and less justified within the broader timeline of the series, reducing its perceived authenticity.

\subsection{LLM Judge Accuracy on Fake and Real Facts}
\label{5.3}

\begin{table}[ht]
\centering
\renewcommand{\arraystretch}{1.4}
\begin{tabular}{|c|>{\centering}p{1.4cm}|>{\centering}p{1.4cm}|>{\centering}p{1.4cm}|>{\centering}p{1.4cm}|>{\columncolor{gray!10}}>{\centering\arraybackslash}p{1.8cm}|}
\hline
\textbf{Judge Model} & \multicolumn{2}{c|}{\textbf{Phi-4}} & \multicolumn{2}{c|}{\textbf{LLaMA-8B}} & \textbf{Real Facts} \\ \hline
\textbf{Plausibility} & \textbf{High} & \textbf{Low} & \textbf{High} & \textbf{Low} & - \\ \hline
\textbf{Falcon-40B} & 89.36 & 92.58 & 86.50 & 91.22 & {32.31} \\ \hline
\textbf{Qwen-72B}   & 59.66 & 57.89 & 53.36 & 56.90 & 80.17  \\ \hline
\textbf{LLaMA-70B}  & 59.66 & 68.34 & 67.29 & 71.81 & 68.03  \\ \hline
\end{tabular}
\vspace{1em}
\caption{\small{Detection accuracy (\%) of each LLM-based judge on fake facts generated by Phi-4 and Llama-8b (under high and low plausibility settings), along with real facts.}}
\label{tab_detection}
\end{table}


We analyze the detection performance of the three LLM judges (Falcon-40b, Llama-70b, and Qwen-72b) in identifying fake information generated by the Phi-4 and Llama-8b models. In particular, Tables \ref{tab_detection} reports the detection accuracy (averaged across all the categories addressed the KG) across both generated fake facts and also real facts. In particular, real facts have been extracted focusing on the same 25 categories discussed above and were generated via Phi-4 using the pipeline described in Section \ref{m:2}. 
Note that, including real facts in the evaluation enables us to assess not only the model's ability to detect fake information, but also its reliability when classifying true statements. This analysis is important for identifying potential detection biases or, more in general, low detection performance, and understanding whether the LLM-based judges are trustworthy even in the presence of factual content.

The results reveal that detection accuracy varies significantly across models. Notably, Falcon-40b demonstrates strong performance in detecting fake information but struggles considerably when evaluating real facts, often misclassifying them. In contrast, Qwen-72b and Llama-70b, despite being large-scale models, show substantially lower accuracy in detecting fake content but perform better in classifying real information.

From these observations, several insights emerge. First, the evaluated LLMs show considerable limitations in accurately distinguishing between fake and real facts. While larger models such as Qwen-72b and Llama-70b appear more balanced in their responses, their accuracy in detecting fake content is around 50–60\%, which is close to random guessing. On the other hand, Falcon-40b, though effective in flagging fake information, tends to adopt an overly conservative stance, labeling most inputs as fake regardless of their actual truthfulness. This behavior suggests a possible bias introduced during training and indicates that Falcon-40b may favor caution over nuance.

These findings underscore the broader challenge of building reliable LLM-based fact verification systems. They also point to the need for future work on improving detection accuracy through targeted fine-tuning, prompt engineering, or hybrid approaches that integrate structured knowledge. Understanding the training dynamics and decision biases of current models is essential for advancing trustworthy and effective detection tools.

\subsection{Category-Wise Analysis of Fake Fact Detectability}
\label{5.2}

\begin{figure}[ht]
    \centering
    \begin{minipage}[b]{0.99\textwidth}
        \centering
        \resizebox{\linewidth}{!}{\includegraphics{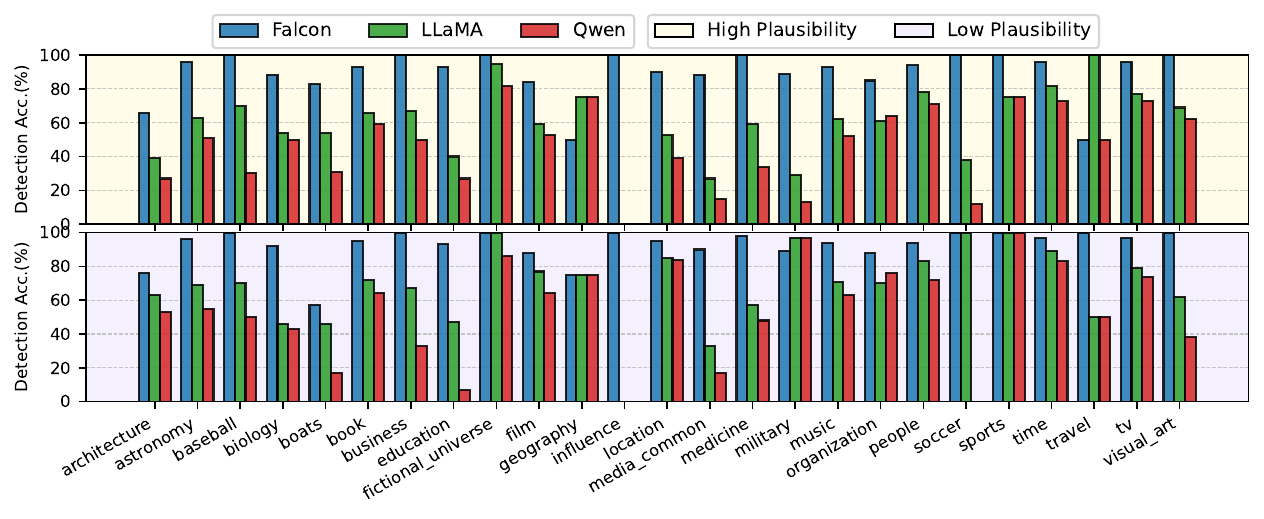}} 
        \vspace{-2em}
        \subcaption{
        \small Fake detector Phi-4.}
        \label{fig:phi4}
    \end{minipage}%
    \hfill
    \begin{minipage}[b]{0.99\textwidth}
        \centering
        \resizebox{\linewidth}{!}{\includegraphics{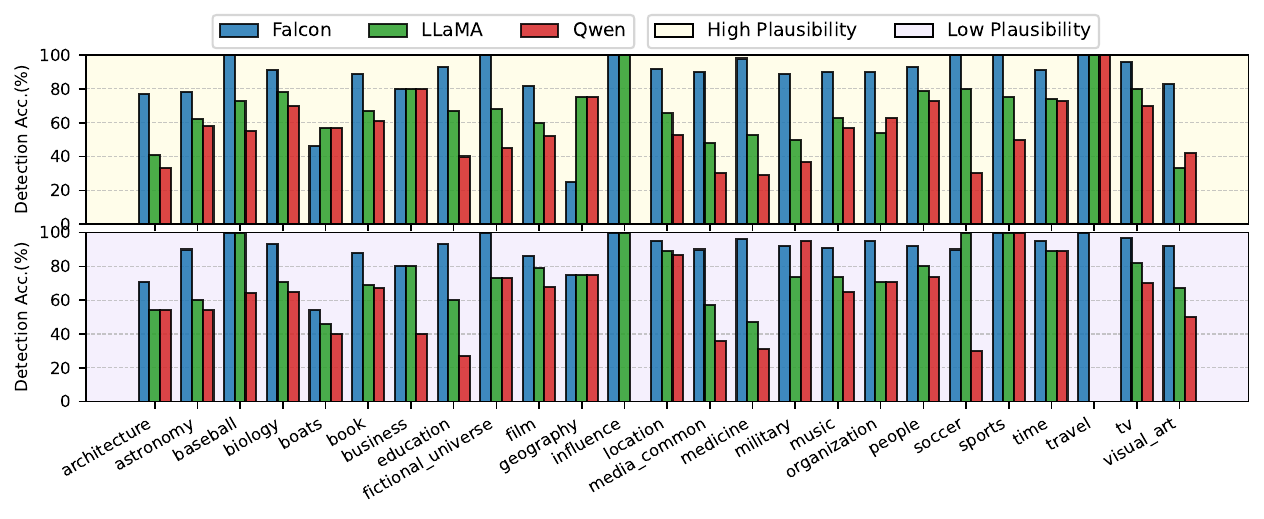}} 
        \vspace{-2em}
        \subcaption{
        \small Fake detector Llama-8b.}
        \label{fig:llama8b}
    \end{minipage}%
    \caption{\small{Detection accuracy across different categories from the knowledge graph. Three LLM-based judges were evaluated: Falcon-40b, Llama-70b, and Qwen-72b (see legend for colors). Fake facts are generated using two models: Phi-4 (top plot) and Llama-8b (bottom plot). Results are shown for fake facts with high plausibility (yellow background) and low plausibility (purple background).}}
    \label{fig:phi-plausibility-analysis}
\end{figure}

This set of experiments investigates the detectability of fake facts across different KG's categories. The analysis has two main objectives: \textit{(i)} to identify potential biases in the detection performance of LLM-based judges across specific categories, and \textit{(ii)} to evaluate whether certain LLMs used for generating fake facts are more capable of producing stealthier, harder-to-detect statements depending on the targeted category.
Note that this analysis focuses exclusively on fake news, with the aim of also understanding how the choice of generative LLM affects the creation of misleading content.

Figure~\ref{fig:phi-plausibility-analysis} presents results across 25 categories. Each subplot distinguishes two levels of plausibility (see Section~\ref{fakeTriplets}): high plausibility is shown with a yellow background, and low plausibility with a purple one. We compare fake information generated by two different LLMs: Phi-4 (top plot) and Llama-8b (bottom plot).

The overall trend observed in the results supports the average findings presented in Table~\ref{tab_detection}, confirming that Falcon-40b tends to be biased toward classifying sentences as fake. While this may initially appear to be a positive outcome, since the current analysis focuses only on fake facts, previous discussion about Table~\ref{tab_detection} shows that this behavior stems from a conservative bias, as Falcon-40b also frequently labels correct facts as fake, indicating a lack of nuance in its judgment.
Despite this limitation, Figure~\ref{fig:phi-plausibility-analysis} shows that the trend is generally consistent across categories, although some category-specific variations exist, for example, in architecture and boats, where performance deviates more significantly.

In contrast, Llama-70b and Qwen-72b generally underperform and exhibit a wide variance in detection accuracy depending on the category. For instance, Llama-70b performs well on 'sports' and 'fictional universe', but shows poor results in many other categories. This highlights how the addressed LLM judges may be ill-suited for detecting misinformation in specific domains. As a consequence, a category-aware analysis could be used by attackers to identify and exploit weaknesses in these models, improving the success of misinformation designed to bypass automated detection. This open interesting challenges for future investigations. 

The effect of plausibility level in fake fact generation is also evident. In some categories, using a higher plausibility significantly reduces detectability. For example, in the 'sports' category in Figure~\ref{fig:phi-plausibility-analysis}(a), the detection rate for Llama and Qwen drops from nearly 100\% for low plausibility fake facts to around 78\% for high plausibility ones. An even more striking example can be seen in the 'travel' category in Figure~\ref{fig:phi-plausibility-analysis}(b), where detection accuracy drops from 100\% to 0\% when moving from low- to high-plausibility facts, again for both Llama and Qwen. These results further confirm that increasing the semantic plausibility of fake facts effectively reduces their detectability, even by state-of-the-art LLM-based detectors.

\section{Conclusion and Future Directions}
\label{con}
This work presents a novel framework that leverages knowledge graphs and large language models to generate structured misinformation in a controlled and semantically grounded way. The proposed pipeline begins by extracting factual triplets from a knowledge graph, representing verifiable truths, and then perturbs them by replacing the object with an alternative entity guided by a plausibility metric (see Section \ref{m:2}). This modified triplet is then transformed into a fluent natural language statement using an LLM conditioned by a specific prompt.

Furthermore, we also explore whether LLMs, without any specific fine-tuning, can act as detectors of fake content. To this end, we introduce a first baseline that relies on prompting LLMs to assess the veracity of given facts.

Experimental results demonstrate that the proposed pipeline is capable of producing fake content that is both wrongly plausible and challenging to detect, particularly when the manipulated triplets maintain a high degree of semantic plausibility. Specifically, Our evaluation of LLMs as detectors reveals significant variability and notable issues in their performance, with outcomes differing considerably across the evaluated models. For instance, Falcon-40b tends to exhibit a conservative bias, frequently labeling sentences as fake, even when they are in fact real. In contrast, models such as Llama-70b and Qwen-72b produce more inconsistent results overall, but show fewer signs of such conservative bias in their predictions.
A more fine-grained analysis indicates that detection effectiveness can also depend on the specific semantic category of the fake information, unveiling category-level weaknesses that can be exploited to further bypass detection.

These findings open several interesting directions for future research. From the detection side, improving LLM-based detection could involve targeted fine-tuning or prompt engineering strategies. From the attack perspective, future work could explore category-aware misinformation generation or adversarial prompting to further evade automated detectors.
Finally, this work highlights the potential of using knowledge graphs as a structured foundation for generating deceptive content, offering a systematic approach to crafting realistic misinformation. It also underscores the dual role of LLMs, both as powerful generators of human-like text and as emerging tools for fact verification. Together, these insights pave the way for future efforts at the intersection of knowledge-based generation and trustworthy AI.

\bibliographystyle{plainnat}
\bibliography{main}
%




\end{document}